\documentclass[letterpaper]{article}
\usepackage{aaai2026}  
\usepackage{times}
\usepackage{helvet}
\usepackage{courier}
\usepackage[hyphens]{url}
\usepackage{graphicx}
\usepackage{natbib}
\usepackage{caption}
\usepackage{amsmath}
\usepackage{footmisc}
\usepackage{amssymb}

\title{Governing Automated Strategic Intelligence}
\newcommand{\primaryauthor}{\textsuperscript{\rm †}}
\newcommand{\equalcontrib}{\textsuperscript{\rm *}}

\author{
    Nicholas~Kruus\equalcontrib\primaryauthor\textsuperscript{\rm 1,2,3},
    Madhavendra~Thakur\equalcontrib\primaryauthor\textsuperscript{\rm 4},
    Adam~Khoja\primaryauthor\textsuperscript{\rm 5},
    Leonhard~Nagel\primaryauthor\textsuperscript{\rm 6},
    Maximilian~Nicholson\primaryauthor\textsuperscript{\rm 7},
    Abeer~Sharma\primaryauthor\textsuperscript{\rm 8},
    Jason~Hausenloy\primaryauthor\textsuperscript{\rm 6}
    \\ \vskip 2ex
    Alberto~KoTafoya\textsuperscript{\rm 4},
    Aliya~Mukhanova\textsuperscript{\rm 4},
    Alli~Katila-Miikkulainen\textsuperscript{\rm 4},
    Harish~Chandran\textsuperscript{\rm 9},
    Ivan~Zhang\textsuperscript{\rm 4},
    Jessie~Chen\textsuperscript{\rm 4},
    Joel~Raj\textsuperscript{\rm 4},
    Jord~Nguyen\textsuperscript{\rm 10,11,12},
    Lai~Hsien~Hao\textsuperscript{\rm 4},
    Neja~Jayasundara\textsuperscript{\rm 13},
    Soham~Sen\textsuperscript{\rm 4},
    Sophie~Zhang\textsuperscript{\rm 4},
    Ashley\textendash Dora~Kokui~Tamaklo\textsuperscript{\rm 4},
    Bhavya~Thakur\textsuperscript{\rm 4},
    Henry~Close\textsuperscript{\rm 4},
    Janghee~Lee\textsuperscript{\rm 14},
    Nina~Sefton\textsuperscript{\rm 4},
    Raghavendra~Thakur\textsuperscript{\rm 4},
    Shiv~Munagala\textsuperscript{\rm 4},
    Yeeun~Kim\textsuperscript{\rm 4}
}
\affiliations{
    \textsuperscript{\rm 1}Schelling Research\\
    \textsuperscript{\rm 2}University of Oxford\\
    \textsuperscript{\rm 3}Ellison Institute of Technology\\
    \textsuperscript{\rm 4}Independent\\
    \textsuperscript{\rm 5}Center for AI Safety\\
    \textsuperscript{\rm 6}ETH Z\"urich\\
    \textsuperscript{\rm 7}University of Bath\\
    \textsuperscript{\rm 8}University of Hong Kong\\
    \textsuperscript{\rm 9}Duke University\\
    \textsuperscript{\rm 10}AI Safety Vietnam\\
    \textsuperscript{\rm 11}Hanoi AI Safety Network\\
    \textsuperscript{\rm 12}University of Science and Technology of Hanoi\\
    \textsuperscript{\rm 13}University of Maine\\
    \textsuperscript{\rm 14}Cornell University\\
    nic@schellingresearch.com, mt3890@columbia.edu\\[0.5ex]
    \primaryauthor Primary authors\quad
}

\begin{document}

\maketitle

\begin{abstract}
Military and economic strategic competitiveness between nation-states will increasingly be defined by the capability and cost of their frontier artificial intelligence models. Among the first areas of geopolitical advantage granted by such systems will be in \textit{automating military intelligence}. Much discussion has been devoted to AI systems enabling new military modalities, such as lethal autonomous weapons, or making strategic decisions. However, the ability of a country of “CIA analysts in a data-center” to \textit{synthesize diverse data at scale}, and its implications, have been underexplored. Multimodal foundation models appear on track to automate strategic analysis previously done by humans. They will be able to fuse today’s abundant satellite imagery, phone-location traces, social media records, and written documents into a single queryable system. We conduct a preliminary uplift study to empirically evaluate these capabilities, then propose a taxonomy of the \textit{kinds} of ground truth questions these systems will answer, present a high-level model of the determinants of this system’s AI capabilities, and provide recommendations for nation-states to remain strategically competitive within the new paradigm of \textit{automated intelligence}.

\end{abstract}


\section{Introduction}

\begin{figure}[!ht]
\centering
\includegraphics[width=0.45\textwidth]{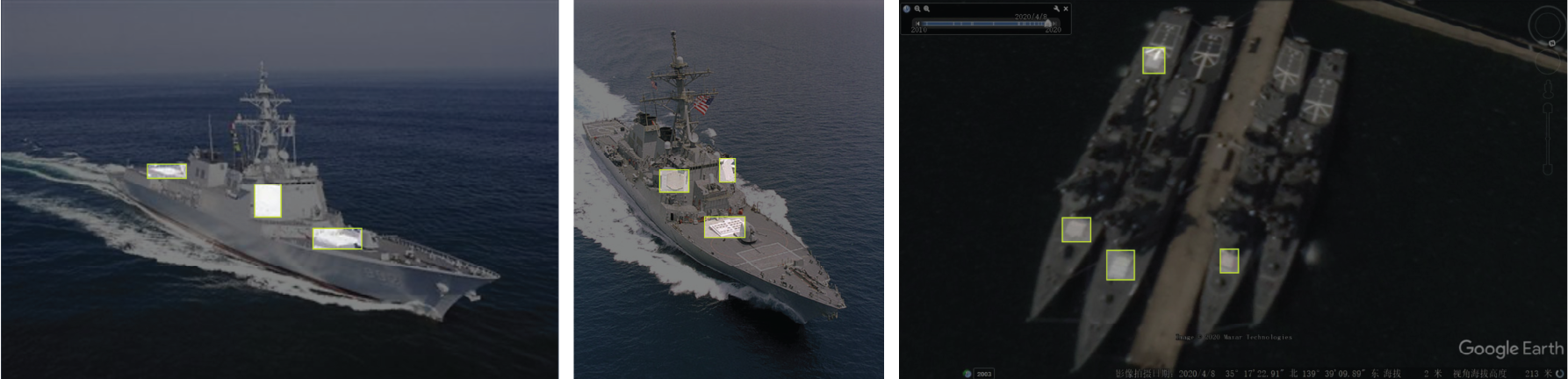}
\caption{Images from the Zhousidun dataset featuring various naval vessels with bounding boxes over SPY radars.}
\label{fig:zhousidun_boxes}
\begin{minipage}{\linewidth}
\footnotesize \textit{Source:} Adapted from \citet{Gupta2024}.
\end{minipage}
\end{figure}

In 2023, researchers discovered that an institute affiliated with the Chinese People’s Liberation Army mistakenly publicized a sensitive dataset of US naval vessels with AI-labeled bounding boxes \cite{Gupta2024}. This is one of countless examples of publicly available open‑source intelligence (OSINT) revealing critical strategic information—that year, an estimated 80-90\% of intelligence analyses in Western countries relied primarily on open sources \cite{Ghioni2024a}. Satellite constellations capture hundreds of terabytes each day \cite{ustin2024earth}, social platforms publish billions of new posts each day \cite{jones2023social,kemp2025digital}, and bulk signals‑intelligence programs record millions of packet headers per second \cite{ghasemirahni2024fajita}.

In 2024 alone, OSINT revealed North‑Korean sanctions‑evasion transfer routes \cite{Salisbury2024}, a Russian missile strike on Kyiv’s Okhmatdyt Hospital \cite{Sheldon2024}, a Chinese prototype reactor for nuclear‑powered aircraft carriers \cite{RisingTang2024}, and much more \cite{Ratcliffe2024,Panella2024,Cabrera2024,Dempsey2024}. Public data—from satellite imagery to social media—are only becoming more important to strategic analysis \cite{USIC2024,ODNI2024,DOD2024}, yet they have already grown more voluminous than human analysts can handle productively \cite{Abadicio2019}.

The data explosion is not confined to OSINT \cite{Koch2024,Christie2020,Farina2014,Whitney2025,NATO1996,WeinerShortND,Wei2018,Tsanousa2022}. For instance, Joint All‑Domain Command and Control aims to connect sensors from \emph{all} US military services—from sea to space—into a single network \cite{CRS2022}.

\subsection{AI Capabilities}
As multimodal AI systems, especially large language models (LLMs), become increasingly common and capable, they are being deployed to address this challenge \cite{Yue2023,Maslej2025,Nyhan2024}. With effective scaffolding, AI systems could fuse satellite imagery, social‑media streams, shipping manifests, and corporate filings, drafting near‑real‑time answers to complex queries at a fraction of a human analyst’s cost \cite{cheng-etal-2023-gpt,zhang2023meta,sustainability_directory_ai_esg_2025,blacksky_geoint_2023,blacksky_platform_2016}. We term this process \textbf{automated intelligence} (\textbf{AUTOINT}). After defining the concept, we outline a five‑stage synthesis pipeline, illustrate six question archetypes, and conclude with policy recommendations.

\subsection{The Burgeoning of Automated Intelligence}
Many believe AI is poised to fundamentally alter the nature of war and intelligence \cite{Black2024,Hendrycks2025,Jahankhani2020,Deloitte2020,EganRosenbach2023,SimmonsEdler2024,Johnson2021,UKMOD2024,McMahon2024,WhiteHouse2025,Jensen2023,GCHQ2021,Koch2024b,Christie2020b,Whitney2025b}. Intelligence agencies worldwide are scrambling to leverage AI, and many have deployed or are developing specialized tools \cite{IARPA2021,DHS2023,Strout2022,IARPA2022}.

In the US, the intelligence community has investigated AI for decades and is now prioritizing implementation \cite{Moran2023,NSCAI2021}. Project Maven alone aims to double or triple analysts’ output using multimodal AI \cite{Pellerin2017,Mohsin2024}; the DOD is rapidly piloting and scaling language models like NIPRGPT and CamoGPT \cite{DoD2023,CDAO2024}; and the CIA has announced a ChatGPT‑style tool for analysts \cite{MartinManson2023,Hindy2023}.

The US is not alone. Chinese strategists rank intelligence among AI’s most critical military applications \cite{Fedasiuk2020}. China’s intelligence services already use AI to identify foreign officers \cite{Wong2023} and are exploring the use of LLMs like ChatBIT to collect, combine, and analyze intelligence information \cite{NelsonEpstein2022,Cheung2023,PomfretPang2024,Insikt2025a}. Russia, which archives massive troves of documents and social‑media posts, has spent years automating stages of the analytic chain \cite{Insikt2025b,SoldatovBorogan2011}. NATO recently partnered with Planet Labs for AI‑enhanced surveillance \cite{Welch2025}. Powers like Iran, North Korea, India, the UK, Germany, Japan, Israel, and non‑state actors are following suit \cite{IranIntl2025,SpecialEurasia2025,GolemND,Haruka2025,LibertadND,Wirtschafter2024}. The race to automate intelligence, in short, is on.

\section{Additional Related Work}
The trend—intelligence data growing faster than analysts’ capacity and surging demand for automation—remains under‑explored. Most research focuses on non-intelligence military applications, such as autonomous kinetic and cyber attacks \cite{AponteGarcia2025} and automated strategic decision-making \cite{Rivera2024}. Existing literature on AI-powered data synthesis lacks typologies and policy implications. While some work measures AI’s analytic aptitude, none evaluate whether AI can \emph{uplift} novices to expert‑level intelligence analysis.

Ghioni and colleagues identify AI as the “informational pivot of intelligence analysis” but do not develop a full account of AUTOINT \cite{Ghioni2024b}. We bridge this gap with our six‑archetype query typology, five‑stage pipeline, and a suite of policy recommendations.

Most studies of AI for intelligence overlook the strategic bottleneck we highlight: \emph{inference} performance \cite{Ghioni2024a}. Training, development, and inference have distinct technical characteristics and political ramifications \cite{Dong2023,Sharma2025,Adams2025,Barros2025}. For instance, bottlenecked inference may exacerbate disparities in access to AUTOINT, considering its tiered pricing  \cite{OpenAIpricing,AnthropicPricing}.
Our preliminary uplift study adapts experimental methods from biosecurity \cite{NASEM2025}. Prior work shows that \emph{experts}’ workflows improve when collaborating with \emph{specialized} AI tools \cite{Toniolo2023} but has not compared \emph{novices} with \emph{publicly available} LLM assistance to analysts.

\section{Exploratory Experiment}
No public dataset is suitable for comparing the intelligence analysis abilities of skilled intelligence analysts to those of novices with and without LLM assistance. In light of this, from June 22 to July 22 this year, we conducted a preliminary study to explore the effects of publicly available LLM use on novice intelligence analysis. In it, 20 novices and 2 skilled analysts addressed 12 intelligence questions.

\paragraph{Procedure} All novices and analysts had the same amount of time (three hours) to answer each intelligence question. In the first round, we randomly assigned participants to either the control group or the LLM-assisted group. LLM usage was prohibited for the control group and required for the LLM-assisted group. Then, we paired participants by experience level to ensure a roughly equal distribution of experience between groups, and we randomly assigned one of the 12 intelligence questions to each pair. We used the same procedure to re-assign pairs, groups, and questions for novices' second question. Participants filled out a spreadsheet as they answered each question. We created three versions of the spreadsheet: one for skilled analysts, one for control novices, and one for LLM-assisted novices.

\paragraph{Questions} The 12 intelligence questions spanned a wide range of topics, data sources, and difficulty levels (see Appendix A for the full list), though we estimated that a professional intelligence analyst could likely reach an accurate conclusion on each one in three hours. Of the novices, 7 were assigned 1 question, while 12 were assigned 2, increasing the study's sample of novice answers to 32. One skilled analyst answered all 12 questions, while the other answered 4. We categorized each question into one of two categories for evaluation: numeric or conceptual. Numeric questions elicited either counts or percentages as answers, while conceptual questions prompted broader and more qualitative responses. See the following two example questions:
\begin{enumerate}
\item \textbf{Numeric:} Between 6/23/2025 and 6/29/2025 (inclusive), what percentage of Starlink terminals shipped to Ukraine showed active signals near the frontline?
\item \textbf{Conceptual:} Map the board connections between ByteDance and state-owned enterprises via public filings.
\end{enumerate}

Since we designed our experiment to accurately reflect real-world intelligence investigations, the questions have no obvious or verifiably correct answer. If participants could directly find accurate answers to our questions online, they could report these answers without conducting any meaningful \emph{analysis}, the primary skill we aim to measure. As a result, we use \emph{similarity to analyst-derived} answers as a proxy for analytical quality. If an intervention makes novices' answers more similar to those of skilled analysts, we consider this notable; it narrows the gap between novice and expert.

\paragraph{Numeric Similarity} For numeric questions, we use an LLM extractor, Gemini 2.5 pro, to identify and extract the numeric answer each participant and expert gave. These identifications required contextual cues for which simpler approaches could not sufficiently account. We chose Gemini 2.5 Pro due to its state-of-the-art performance \cite{comanici2025gemini25}. A random sample of ~50\% of Gemini's extractions were human-validated, with no errors found. To compare the extracted numeric answers from participants to those from analysts, we use a symmetric log-ratio similarity with an offset safe for zero or near-zero values. Let $a$ denote a participant’s extracted value and $r$ the per-question expert reference (percents are normalized to fractions during extraction). We define a scale-aware continuity term
\[\tau=\max\{\epsilon_{\mathrm{abs}}(q),\,\epsilon_{\mathrm{rel}}\lvert r\rvert\},\]
where $\epsilon_{\mathrm{rel}}=10^{-6}$ and $\epsilon_{\mathrm{abs}}(q)$ is a small absolute floor that depends on question type (e.g., $10^{-6}$ for proportions, $0.5$ for counts). The log-ratio distance is \[d=\bigl\lvert\log\!\frac{a+\tau}{r+\tau}\bigr\rvert,\]
and we convert it to a similarity $S=\exp(-d)\in(0,1]$. This is (i) \emph{scale invariant} (multiplying both $a$ and $r$ by a constant leaves $S$ unchanged), (ii) \emph{symmetric} around the reference (over‐ and underestimation by the same factor incur equal penalties), (iii) \emph{zero safe} without arbitrary clipping, as $\tau$ anchors continuity to the expert scale, and (iv) it preserves \emph{order-of-magnitude resolution} (e.g., $10\times$ vs.\ $100\times$ differences yield clearly separated similarities) rather than saturating as in sMAPE or normalized absolute errors. Moreover, measuring error in log space aligns with multiplicative noise common in counts and rates, stabilizing variance and facilitating aggregation; $\epsilon_{\mathrm{rel}}$ and $\epsilon_{\mathrm{abs}}(q)$ provide transparent, auditable knobs without altering the core invariance and symmetry properties.

\paragraph{Conceptual Similarity} For conceptual questions, an Gemini 2.5 Pro examined the \emph{semantic} similarity between participant and analyst responses to conceptual questions. In context, we provided Gemini instructions, a participant's response to a given question, and the analyst response to the same question. It was blinded to the response condition (control or LLM-assisted). We required LLM evaluations to use a 1–5 point scale, following evidence that the correlation between LLM scores on such a scale and human ones ($r = 0.417$) is comparable to that produced by a less feasible state-of-the-art approach ($r = 0.514$) \cite{liu-etal-2023-g}. These scores were normalized to a $[0, 1]$ scale before being analyzed alongside the numeric similarity scores above.

\subsection{Statistical Approach}
We employ one primary statistical technique alongside two others, acting as robustness checks.

\paragraph{Cluster-Robust Regression with Fixed Effects} Our primary estimate of the LLM effect uses an ordinary least squares (OLS) regression with item and metric fixed effects and cluster-robust inference at the participant level. This specification uses all observations and is identified primarily by differences between participants and secondarily by changes within participants where present. (Some participants were assigned to both groups due to their participation in both rounds of the study.) We report dedicated within-participant/matched-pairs estimates as robustness checks. Let $y_{imq}\in[0,1]$ denote the (normalized) outcome for participant $i$, metric type $m$, and question $q$, and let $T_{imq}\in\{0,1\}$ indicate assignment to the LLM-assisted group. The primary specification is
\[
y_{imq} \;=\; \alpha \;+\; \beta\,T_{imq} \;+\; \gamma_m \;+\; \delta_q \;+\; \varepsilon_{imq},
\]
where $\gamma_m$ and $\delta_q$ are fixed effects for the type of metric (numeric or conceptual) and the question, respectively. We fit this model via OLS and compute Huber-White cluster-robust standard errors clustered on participant $i$ (with the small-sample finite-sample correction provided by the estimator). Hypothesis tests target a directional alternative, $H_0\!:\beta\le 0$ vs.\ $H_1\!:\beta>0$, using a one-sided $z$-statistic derived from the cluster-robust estimate of $\beta$'s variance. We use this directional test because it reflects the purpose of this study: to examine whether LLMs narrow the gap between novice and analyst performance. The broader approach (i) absorbs systematic differences in baseline difficulty or scaling across questions and metrics through $\gamma_m$ and $\delta_q$, preventing item/metric composition from confounding the between-group contrast; and (ii) accounts for within-participant dependence and heteroskedasticity via clustering, which is essential because each participant contributes multiple observations across questions/metrics. For interpretation, the estimated between-group effect is the adjusted mean difference $\hat{\beta}$; equivalently, adjusted marginal means are obtained by predicting each observation under $T=1$ and $T=0$ and averaging to yield $\hat{\mu}_T$ and $\hat{\mu}_C$, with $\hat{\beta}=\hat{\mu}_T-\hat{\mu}_C$.

\paragraph{Stratified Permutation} As a robustness check for any between-group effects, we estimate a finite-sample valid $p$-value by performing a stratified permutation test at the participant-by-period level. Let $i$ index participants (clusters) and $t$ index randomized periods (waves/sends). We map observed group labels to a binary treatment $T_{it}\in\{0,1\}$ (discarding extraneous categories) and enforce a single, consistent label within each $(i,t)$ cell before analysis. The observed test statistic $\hat{\tau}$ is a pre-specified difference in means $\bar{Y}_{T=1}-\bar{Y}_{T=0}$ for the outcome of interest (e.g., residual or score); optionally, we collapse to cluster means to equal-weight participants and limit leverage from high-volume users. We then generate $B=10,000$ permutations by \emph{relabeling treatment within each period} $t$ across participants $i$, preserving the treated count per stratum, and recompute $\hat{\tau}^{(b)}$; the one-sided randomization $p$-value for $H_1:\text{Treatment}>\text{Control}$ is \[p=\big(1+\sum_{b=1}^{B}\mathbb{I}\{\hat{\tau}^{(b)}\ge \hat{\tau}\}\big)/(B+1).\] This design-based procedure is justified because it (i) aligns the null distribution with the actual assignment mechanism (exchangeability \emph{within} period), (ii) provides exact validity under the sharp null without large-sample or parametric assumptions, (iii) respects clustering and permits equal-weighting to mitigate imbalance, and (iv) avoids spurious significance that can arise from model misspecification or heteroskedasticity. When no period variable exists, we conservatively drop clusters exhibiting mixed labels and permute a single label per cluster, ensuring the null remains well-defined.

\paragraph{Residual-Difference}
We estimate the treatment effect using a paired residual–difference test that is explicitly aligned with the crossover design and with the way assignment was implemented. Let \(Y_{imqs}\) denote the outcome for participant \(i\), metric \(m\), question \(q\), and study \(s\). To remove systematic nuisance variation that does not reflect the treatment effect, we first residualize \(Y_{imqs}\) on metric, question, and study fixed effects via OLS, \[Y_{imqs}=\alpha+\mu_m+\gamma_q+\sigma_s+\varepsilon_{imqs},\] and define \(\hat r_{imqs}=\hat\varepsilon_{imqs}\). We then restrict to “crossover’’ participants who appear in both arms \(a\in\{\text{Control},\text{Treatment}\}\), compute per–arm mean residuals \(\bar r_{ia}\), and form the within–participant difference \(d_i=\bar r_{i,\text{Treatment}}-\bar r_{i,\text{Control}}\). The test statistic \(t=\bar d/(s_d/\sqrt{n})\) with \(n-1\) degrees of freedom, where \(\bar d\) and \(s_d^2\) are the sample mean and variance of \(\{d_i\}_{i=1}^n\), targets \(H_0:\mathbb{E}[d_i]\le 0\) versus the preregistered directional alternative \(H_1:\mathbb{E}[d_i]>0\), and we report the one–sided \(p=1-F_{t_{n-1}}(t)\) and the one–sided \((1-\alpha)\) lower confidence bound \(\bar d - t_{1-\alpha,n-1}s_d/\sqrt{n}\). This approach is methodologically defensible for three reasons: (i) it conditions on all nonrandom, design–level heterogeneity (metric, question, study) without imposing cross–arm equality of those effects, thereby reducing bias and variance; (ii) it preserves the paired nature of the design and collapses to one contrast per participant, which eliminates the need for cluster–robust variance corrections and gives each participant equal weight in the estimand; and (iii) it matches the actual randomization scheme—when assignment is constant within a study block, within–block pairing is undefined or low–power, whereas residualizing by study fixed effects absorbs between–study shifts while retaining all crossover information. Relative to mixed–effects or cluster–robust regressions on the raw panel, the paired residual–difference test makes weaker assumptions about the error structure, is transparent, and directly estimates for Treatment \(>\) Control.

\subsection{Participant Demographics}
We asked participants optional demographic questions. The mean self-reported hours of intelligence analysis experience for skilled analysts (350.000) was over 9 times greater than that for novices (38.800). The control and LLM-assisted groups had 39.375 and 32.727 mean hours of experience, respectively. All novices and skilled analysts were volunteers fluent in English, the language of the experiment's materials. The male-to-female sex ratio was three to two.

In a separate optional survey, we asked LLM-assisted participants about their AI usage habits. Six of the ten respondents used AI models daily before the study, three used them weekly, and one used them monthly. We gave control participants a placebo survey on their research habits.

When asked in a post-study free-response question, only one respondent correctly identified the study's purpose, suggesting that demand characteristics did not affect our results.

\subsection{Results}
In the \textit{primary} analysis, we find an LLM effect on the unified 0--1 expert-similarity score of \(\hat\beta=0.148\) with \(\mathrm{SE}=0.0606\), yielding \(t=2.44\) (\(\mathrm{df}=4\)) and \(p=0.0355\) for the directional hypothesis $\texttt{LLM-assisted}>\texttt{control}$. Two robustness checks align in direction and one aligns in significance: a paired test on residualized participant-level means gives \(\Delta=0.153\), \(t=2.796\), one-sided \(p=0.025\); and a participant-stratified permutation test on residual mean differences yields \(\Delta=0.086\) with one-sided \(p=0.219\).

These results indicate that the LLM-assisted novices' responses were significantly more similar to those of skilled analysts than the responses of novices without LLM assistance.
Despite the lack of complete robustness in our results, we believe these findings warrant serious attention to and discussion of how LLMs, even without fine-tuning, may democratize high-quality intelligence analysis and what it implies broadly.

\section{The Five‑Stage Synthesis Pipeline}\label{sec:pipeline}

The automated conversion of raw data into actionable intelligence analysis might follow a five-stage pipeline:

\begin{enumerate}
    \item \textbf{Ingestion:} Gather and deduplicate relevant feeds, such as images, broadcasts, text, and tables.
    \item \textbf{Representation:} Convert every item into formats (vector embeddings, written summaries, etc.) amenable to tool-based agentic queries.
    \item \textbf{Retrieval:} Translate AI agent tool calls into database queries that return specific and relevant information.
    \item \textbf{Reasoning:} Apply judgmental and quantitative analysis over retrieved data, informing further tool calls and ultimately producing answers with confidence scores.
    \item \textbf{Integration:} Send analysis to decision-makers, paired with provenance and confidence information. Record it for later reference by other AI or human analysts.
\end{enumerate}

\subsection{Illustrative Scenario}

Consider a frontier multimodal model licensed through a classified partnership with a leading AI lab, fine‑tuned extensively on historical intelligence briefings and deployed within a secure, air‑gapped enclave \citep{Mitchell2025GPT4oTopSecret,Anthropic2025DoDOTA}. %
This model differs significantly from commercial variants—it has undergone specialized reinforcement learning (RL) from analyst feedback and ground-truth results of thousands of relevant strategic assessments and quantitative forecasts, giving it domain‑specific fluency unmatched by public systems \citep{christiano2017deep,ouyang2022training,anthropic2025claudegov}. %
The model maintains persistent tool‑use connections to real‑time satellite imagery, signals‑intelligence repositories, weather forecasts, and previous intelligence reports \citep{schick2023toolformer,guu2020realm}. Its RL training was conducted with these tools ``in-the-loop,'' giving it an intuitive understanding of when and how to query relevant information \cite{deepresearch,openai2025o3,nakano2021webgpt,jin2025searchr1}.

The task of evaluating ambush risk along two potential extraction routes for a high‑value asset in contested territory illustrates the full analysis pipeline.
An instance of the AI model tasked with this evaluation queries for information on the relevant geographic corridors and time‑frames. The queries are run against gigabytes of fresh preprocessed data—the latest satellite passes over the region, intercepted communications, weather patterns affecting visibility and mobility, and known troop movements. %
The model spins up and delegates auxiliary tasks to subagents, fluidly considering and integrating subagents' specialized assessents into its broader understanding of the situation \citep{hadfield2025multiagent,schroeder-etal-2025-thread,zhu-etal-2024-redel}.
Next, the model reasons through adversary actions, terrain constraints, and historical ambush patterns, generating exposure‑risk scores for each route segment \citep{ownby_kott_2006_raid,geng2020deep}.

Within twenty minutes, a structured assessment is complete: Route B offers 63\% lower ambush risk than Route A in expectation, based primarily on recent changes in local patrol patterns and satellite‑detected brush‑clearing activities. %
The assessment includes annotated map overlays, supporting‑evidence snippets, and confidence intervals—all formatted for immediate integration into command briefings.
Military leadership, while retaining ultimate authority, now bases their extraction plan on a comprehensive analysis that would have taken far more time for humans to produce. %
The result is faster operational tempo, reduced intelligence blind spots, and higher success rates.


\section{Political Implications}

Automated intelligence capabilities fundamentally alter the strategic landscape of national security and intelligence operations. They present opportunities and risks that demand immediate, global policy attention.

\subsection{Geopolitical Ramifications}

AUTOINT systems pose three primary challenges to existing intelligence hierarchies. First, they enable smaller states and non-state actors to achieve near-parity with established powers in analytical capabilities, effectively nullifying the intelligence arbitrage that status-quo powers have long relied upon for strategic advantage \citep{Kreps2021NonstateAI}. A nation with limited human intelligence resources but access to frontier AI models may now be able to process and synthesize open-source data at scales previously reserved for major intelligence agencies \citep{Ghioni2024OSINTAI}.

Second, the democratization of analytical capabilities may increase asymmetric threats, particularly terrorism. While terrorist organizations typically possess sufficient destructive capabilities, one of their main operational bottlenecks has historically been logistical coordination and intelligence synthesis \cite{tsvetovat2003structural}. AUTOINT removes this constraint, potentially enabling more sophisticated and coordinated attacks by providing automated operational planning and target identification capabilities \citep{UNICRI2021AlgorithmsTerrorism}.

Third, as analytical processes become increasingly automated, intelligence superiority will be increasingly determined by access to proprietary data and advance models, not human expertise or capacity. This shift will drive intelligence organizations to refocus from data analysis toward data acquisition, potentially intensifying espionage activities, cyber operations, and other collection methods \citep{NSCAI2021FinalReport}. The resulting competition for exclusive data sources may destabilize existing intelligence-sharing agreements and other cooperations.

\subsection{Strategic Vulnerabilities}

The integration of AI systems into intelligence operations introduces novel attack vectors that adversaries may exploit. AUTOINT systems become high-value targets for manipulation, requiring robust defenses against adversarial inputs, model poisoning, and prompt injection attacks \citep{NIST_AI_100_2e2025}. The reliability and alignment of these systems become critical national security concerns, as compromised or misaligned AI could provide adversaries with strategic advantages or generate misleading intelligence that undermines decision-making.

The concentration of analytical capabilities in AI systems may also create single points of failure. Unlike distributed human analysts, centralized AUTOINT systems may be more vulnerable to targeted cyber attacks or technical failures that could cripple domestic intelligence capacity.

\subsection{Policy Recommendations}

To maintain strategic competitiveness in the AUTOINT era while mitigating its risks, we recommend governments pursue a suite of policies focused chiefly on five areas.

\subsubsection{AI Infrastructure Protection}

Governments should implement protectionist policies for critical AI infrastructure, including compute resources, specialized hardware, and foundational models. This includes export controls on advanced semiconductors and AI accelerators to prevent adversaries from developing superior AUTOINT capabilities \citep{BIS_FedReg_AdvancedComputing_2023}. Additionally, nations should consider mandating the expulsion of foreign-made components from domestic AI infrastructure to reduce supply chain vulnerabilities and potential backdoors \citep{CRS_IN11663_2025}. Complementary to these defensive measures, governments should incentivize domestic semiconductor and AI development through targeted subsidies, research grants, and public-private partnerships \citep{WhiteHouse_CHIPS_FactSheet_2022}. Building indigenous AI capabilities reduces dependence on foreign technologies and ensures continued access to frontier models even under adverse geopolitical conditions.

\subsubsection{Data Sovereignty and Security}

Nations must establish comprehensive data sovereignty frameworks ensuring that sensitive information remains under domestic control \citep{DOJ2025_EO14117_FinalRule}. This requires legislation mandating that critical datasets—including government records, infrastructure data, and citizen information—be stored and processed onshore \citep{DOJ2025_EO14117_FinalRule}. Cross-border data flows should be subject to strict controls, particularly for information that adversaries could leverage via AUTOINT \citep{RAND_2024_BulkDataAI}. Governments should also implement robust information and cybersecurity measures to prevent advanced AUTOINT systems and their insights from being compromised through cyber attacks, insider threats, or technical vulnerabilities \citep{RAND_2024_BulkDataAI,NIST_SP_800_207_2020}. This includes developing secure computing environments, implementing zero-trust architectures, and establishing incident response protocols specifically designed to handle AI systems \citep{NIST_SP_800_207_2020,NIST_AIRMF_1_0_2023}.

\subsubsection{Open Source Intelligence Management}

The enhanced synthesis capabilities of AUTOINT systems necessitate a fundamental reassessment of open source information security \citep{OHCHR_Berkeley_Protocol_2022}. Nations should conduct comprehensive audits of their open source footprints, identifying publicly available information that could be aggregated and analyzed by adversarial AUTOINT systems to reveal sensitive intelligence \citep{OHCHR_Berkeley_Protocol_2022,RAND_2024_BulkDataAI}. Policy measures should include guidelines for government agencies and critical infrastructure operators on limiting sensitive information disclosure in public forums, social media, and official publications \citep{OHCHR_Berkeley_Protocol_2022}. This may require revising freedom of information laws and public disclosure requirements to account for the enhanced analytical capabilities that AI provides to potential adversaries \citep{RAND_2024_BulkDataAI}.

\subsubsection{AI Alignment and Reliability}

Given the critical role of AUTOINT systems in national security decision-making, governments must prioritize AI alignment and reliability research. This includes developing standards for AI system verification and validation, establishing testing protocols to ensure analytical outputs are accurate and unbiased, and creating oversight mechanisms for AI-generated intelligence products. Governments should invest in research programs focused on making AI systems more interpretable and reliable, ensuring that intelligence analysts can understand and validate AI-generated conclusions \citep{NIST_AIRMF_1_0_2023}. Additionally, developing robust human-AI collaboration frameworks will help maintain human oversight while leveraging AI capabilities \citep{NIST_AIRMF_1_0_2023,NATO_2024_AI_Strategy_Revised}.

\subsubsection{International Cooperation and Norms}

The global nature of AUTOINT capabilities requires coordinated international responses \citep{NATO_2024_AI_Strategy_Revised}. Governments should work through existing multilateral frameworks to establish norms governing the responsible development and deployment of AI for intelligence purposes. This includes developing agreements on prohibited uses of AUTOINT systems, establishing protocols for preventing AI-enabled intelligence operations that could escalate international tensions, and creating mechanisms for sharing threat intelligence related to AI system vulnerabilities \citep{NATO_2024_AI_Strategy_Revised,RAND_2024_BulkDataAI}.

\subsubsection{Quantitative Benchmarking} 

Establish dedicated evaluation units that continuously measure accuracy, calibration, hallucination rates, and adversarial robustness—drawing on judgmental forecasting and analogous industrial safety audits—to build institutional trust before mission‑critical deployment \cite{toreini2020, park2022, tetlock2015superforecasting}.

\section{Conclusion}

We find evidence that artificial intelligence systems may democratize quality intelligence analysis. This has a host of political implications, motivating our recommendations. Successful implementation of our suggestions requires a careful balance between security and economic considerations. Overly restrictive policies may hinder domestic AI development and economic competitiveness, while insufficient protections could severely disadvantage nations.

Governments should adopt graduated implementation strategies, beginning with the most critical infrastructure and datasets while gradually extending protections across the broader AI ecosystem. Regular assessment and adaptation of policies will be necessary as AI capabilities continue to evolve and new threats emerge.

The transition to AUTOINT-aware governance will require substantial investment in new institutions and technologies. However, the strategic advantages conferred by effective policy implementation—and the risks of inaction—make such investments essential for national security in the AI era.

\appendix
\section{Appendix A: Full Intelligence Question List}
Below we present the full list of intelligence questions that participants and analysts answered (random order):
\begin{enumerate}
    \item Which VKontakte pages operated by or closely linked to Russias GRU have posted about NATO activities from 6/20/2025 to 6/27/2025, and how many posts did each make?
    \item Identify Iranian universities publishing drone research with Chinese co-authors after 2023 and before 6/23/2025.
    \item Profile the academic backgrounds of Huawei's top 20 6G researchers using Linkedin/papers.
    \item Which ship-to-ship transfers of Russian-origin crude oil have occurred within Singapores territorial waters (12 nm) between 5/24/2025 and 6/23/2025 (inclusive of both end dates), according to public AlS data?
    \item How many Russian oligarch yachts were docked in UAE marinas on 6/23/2025 based on AlS data?
    \item As of 6/23/2025, which cryptocurrency exchanges still process transactions linked to North Korean addresses?
    \item How many Turkish firms registered new subsidiaries in Russia after Western companies exited and before 6/23/2025?
    \item Between 6/23/2025 and 6/29/2025 (inclusive of both end dates), what percentage of Starlink terminals shipped to Ukraine showed active signals near the frontline?
    \item Which Singapore free trade zones saw the highest increase in transshipments to Russia post-sanctions?
    \item Which Belgian or Dutch ports handled the most dual-use chemical shipments to Syria in 2024?
    \item Map the board connections between ByteDance and state-owned enterprises using public filings.
    \item Which Chinese Al companies filed US patents in 2024 despite being on the Entity List?
\end{enumerate}

\bibliography{references}

\begin{thebibliography}{123}
\providecommand{\natexlab}[1]{#1}

\bibitem[{Abadicio(2019)}]{Abadicio2019}
Abadicio, F. 2019.
\newblock Data Deluge: Challenges and Opportunities in Intelligence Analysis.
\newblock \emph{Intelligence and National Security}, 34(6): 811--830.

\bibitem[{Adams(2025)}]{Adams2025}
Adams, J. 2025.
\newblock Serving LLMs at Scale: A Systems Perspective.
\newblock \emph{Communications of the ACM}, 68(7): 88--97.

\bibitem[{Anthropic(2025{\natexlab{a}})}]{Anthropic2025DoDOTA}
Anthropic. 2025{\natexlab{a}}.
\newblock Anthropic and the Department of Defense to Advance Responsible AI in Defense Operations.

\bibitem[{Anthropic(2025{\natexlab{b}})}]{AnthropicPricing}
Anthropic. 2025{\natexlab{b}}.
\newblock Claude API Pricing.
\newblock Web Page.
\newblock Accessed 2025-07-07.

\bibitem[{Anthropic(2025{\natexlab{c}})}]{anthropic2025claudegov}
Anthropic. 2025{\natexlab{c}}.
\newblock Claude Gov Models for U.S. National Security Customers.

\bibitem[{Aponte~Garc{\'i}a et~al.(2025)Aponte~Garc{\'i}a, Mart{\'i}nez~Barrios, Romero-S{\'a}nchez, Aponte~Garc{\'i}a, and Garc{\'i}a~Vald{\'e}s}]{AponteGarcia2025}
Aponte~Garc{\'i}a, C.~A.; Mart{\'i}nez~Barrios, H.~E.; Romero-S{\'a}nchez, A.; Aponte~Garc{\'i}a, M.~S.; and Garc{\'i}a~Vald{\'e}s, M. d.~P. 2025.
\newblock Governance and Regulation of Autonomous Weapons and Cybersecurity (2016–2024): The Influence of States, International Organizations, and Civil Society on International Humanitarian Law.
\newblock \emph{Contemporary Readings in Law and Social Justice}, 17(1): 550--562.

\bibitem[{Barros(2025)}]{Barros2025}
Barros, M. 2025.
\newblock Bandwidth Constraints for Remote AI Inference.
\newblock \emph{IEEE Transactions on Networking}, 33(3): 605--617.

\bibitem[{Black(2024)}]{Black2024}
Black, S. 2024.
\newblock AI on the Battlefield: An Overview of Emerging Capabilities.

\bibitem[{{BlackSky}(2016)}]{blacksky_platform_2016}
{BlackSky}. 2016.
\newblock It's here! Introducing the {BlackSky} global intelligence platform.
\newblock Accessed 2025-07-25.

\bibitem[{{BlackSky}(2023)}]{blacksky_geoint_2023}
{BlackSky}. 2023.
\newblock Three {GEOINT} trends spotted at {Esri Federal GIS} 2023.
\newblock Accessed 2025-07-25.

\bibitem[{Cabrera(2024)}]{Cabrera2024}
Cabrera, L. 2024.
\newblock Tracking Russian Armor on TikTok.
\newblock Accessed 2025-07-07.

\bibitem[{Cheng, Li, and Bing(2023)}]{cheng-etal-2023-gpt}
Cheng, L.; Li, X.; and Bing, L. 2023.
\newblock Is {GPT}-4 a Good Data Analyst?
\newblock In Bouamor, H.; Pino, J.; and Bali, K., eds., \emph{Findings of the Association for Computational Linguistics: EMNLP 2023}, 9496--9514. Singapore: Association for Computational Linguistics.

\bibitem[{Cheung(2023)}]{Cheung2023}
Cheung, T.~M. 2023.
\newblock China’s Race for Military AI Dominance.
\newblock \emph{Journal of Strategic Studies}, 46(5): 792--820.

\bibitem[{{Chief Digital and AI Office of the US DOD}(2023)}]{DoD2023}
{Chief Digital and AI Office of the US DOD}. 2023.
\newblock Year in Review 2023.
\newblock Accessed 2025-07-07.

\bibitem[{{Chief Digital and AI Office of the US DOD}(2024)}]{CDAO2024}
{Chief Digital and AI Office of the US DOD}. 2024.
\newblock CamoGPT: DoD’s Secure Large Language Model.
\newblock Accessed 2025-07-07.

\bibitem[{Christiano et~al.(2017)Christiano, Leike, Brown, Martic, Legg, and Amodei}]{christiano2017deep}
Christiano, P.; Leike, J.; Brown, T.; Martic, M.; Legg, S.; and Amodei, D. 2017.
\newblock Deep Reinforcement Learning from Human Preferences.
\newblock In \emph{Advances in Neural Information Processing Systems 30}, 4299--4307.

\bibitem[{Christie(2020{\natexlab{a}})}]{Christie2020}
Christie, R. 2020{\natexlab{a}}.
\newblock The Expanding Sensor Web: Implications for Defense Intelligence.
\newblock \emph{Defense Studies}, 20(3): 225--243.

\bibitem[{Christie(2020{\natexlab{b}})}]{Christie2020b}
Christie, R. 2020{\natexlab{b}}.
\newblock Training vs. Inference: A Technical Separation.
\newblock \emph{AI Magazine}, 41(3): 22--30.

\bibitem[{Comanici et~al.(2025)Comanici, Bieber, Schaekermann, Pasupat, Sachdeva, Dhillon et~al.}]{comanici2025gemini25}
Comanici, G.; Bieber, E.; Schaekermann, M.; Pasupat, I.; Sachdeva, N.; Dhillon, I.; et~al. 2025.
\newblock Gemini 2.5: Pushing the Frontier with Advanced Reasoning, Multimodality, Long Context, and Next Generation Agentic Capabilities.
\newblock \emph{arXiv preprint arXiv:2507.06261}.

\bibitem[{Dempsey(2024)}]{Dempsey2024}
Dempsey, J. 2024.
\newblock Open-Source Sleuths Pinpoint Iranian Drone Factories.
\newblock Accessed 2025-07-07.

\bibitem[{Dong et~al.(2023)}]{Dong2023}
Dong, H.; et~al. 2023.
\newblock Efficient Large-Scale Training of Foundation Models.
\newblock In \emph{Proceedings of the 40th International Conference on Machine Learning}.

\bibitem[{Egan and Rosenbach(2023)}]{EganRosenbach2023}
Egan, P.; and Rosenbach, E. 2023.
\newblock AI and the Changing Character of War.
\newblock \emph{Foreign Affairs}, 102(6): 88--101.

\bibitem[{{Executive Office of the US President}(2025)}]{WhiteHouse2025}
{Executive Office of the US President}. 2025.
\newblock National Strategy for Critical Infrastructure and Artificial Intelligence.
\newblock Accessed 2025-07-07.

\bibitem[{Farina(2014)}]{Farina2014}
Farina, A. 2014.
\newblock Signals Everywhere: The Rise of SIGINT in Open Networks.
\newblock \emph{Journal of Information Warfare}, 13(2): 47--60.

\bibitem[{Fedasiuk(2020)}]{Fedasiuk2020}
Fedasiuk, R. 2020.
\newblock Harnessed Lightning: China’s Military AI Strategy.
\newblock Accessed 2025-07-07.

\bibitem[{Freeman et~al.(2022)Freeman, Koenig, Stover, and {Office of the United Nations High Commissioner for Human Rights}}]{OHCHR_Berkeley_Protocol_2022}
Freeman, L.; Koenig, A.; Stover, E.; and {Office of the United Nations High Commissioner for Human Rights}. 2022.
\newblock \emph{Berkeley Protocol on Digital Open Source Investigations: A Practical Guide on the Effective Use of Digital Open Source Information in International Human Rights, Humanitarian Law and Criminal Investigations}.
\newblock New York: United Nations.

\bibitem[{Gallagher(2025)}]{CRS_IN11663_2025}
Gallagher, J.~C. 2025.
\newblock Secure and Trusted Communications Networks Reimbursement Program: Frequently Asked Questions.
\newblock Technical Report IN11663, Congressional Research Service.

\bibitem[{Geng et~al.(2020)Geng, Liu, Wang, and Liu}]{geng2020deep}
Geng, Y.; Liu, E.; Wang, R.; and Liu, Y. 2020.
\newblock Deep Reinforcement Learning Based Dynamic Route Planning for Minimizing Travel Time.
\newblock \emph{CoRR}, abs/2011.01771.

\bibitem[{Ghasemirahni et~al.(2024)Ghasemirahni, Farshin, Scazzariello, Maguire~Jr., Kosti\'{c}, and Chiesa}]{ghasemirahni2024fajita}
Ghasemirahni, H.; Farshin, A.; Scazzariello, M.; Maguire~Jr., G.~Q.; Kosti\'{c}, D.; and Chiesa, M. 2024.
\newblock FAJITA: Stateful Packet Processing at 100 Million pps.
\newblock \emph{Proc. ACM Netw.}, 2(CoNEXT3).

\bibitem[{Ghioni, Taddeo, and Floridi(2023)}]{Ghioni2024a}
Ghioni, R.; Taddeo, M.; and Floridi, L. 2023.
\newblock Open-Source Intelligence (OSINT) and AI: The Informational Pivot of Intelligence Analysis.
\newblock Oxford Internet Institute Blog.
\newblock Accessed 2025-07-07.

\bibitem[{Ghioni, Taddeo, and Floridi(2024{\natexlab{a}})}]{Ghioni2024b}
Ghioni, R.; Taddeo, M.; and Floridi, L. 2024{\natexlab{a}}.
\newblock AI in Intelligence Analysis: A Systematic Review.
\newblock \emph{AI and Society}, 39(2): 345--368.

\bibitem[{Ghioni, Taddeo, and Floridi(2024{\natexlab{b}})}]{Ghioni2024OSINTAI}
Ghioni, R.; Taddeo, M.; and Floridi, L. 2024{\natexlab{b}}.
\newblock Open source intelligence and AI: a systematic review of the GELSI literature.
\newblock \emph{AI \& Society}, 39: 1827--1842.

\bibitem[{Group(2025{\natexlab{a}})}]{Insikt2025a}
Group, I. 2025{\natexlab{a}}.
\newblock Artificial Eyes: Generative AI in China’s Military Intelligence.
\newblock Accessed 2025-07-07.

\bibitem[{Group(2025{\natexlab{b}})}]{Insikt2025b}
Group, I. 2025{\natexlab{b}}.
\newblock Automating Intelligence: Russia’s AI ambitions.
\newblock Accessed 2025-07-07.

\bibitem[{Gupta et~al.(2024)}]{Gupta2024}
Gupta, S.; et~al. 2024.
\newblock {Zhousidun}: An Open-Source Dataset of US Naval Vessels.
\newblock Blog post, Import AI Newsletter No. 374.
\newblock Accessed 2025-07-07.

\bibitem[{Guu et~al.(2020)Guu, Lee, Tung, Pasupat, and Chang}]{guu2020realm}
Guu, K.; Lee, K.; Tung, Z.; Pasupat, P.; and Chang, M. 2020.
\newblock {REALM}: Retrieval-Augmented Language Model Pre-Training.
\newblock \emph{arXiv preprint arXiv:2002.08909}.

\bibitem[{Hadfield et~al.(2025)Hadfield, Zhang, Lien, Scholz, Fox, and Ford}]{hadfield2025multiagent}
Hadfield, J.; Zhang, B.; Lien, K.; Scholz, F.; Fox, J.; and Ford, D. 2025.
\newblock How we built our multi-agent research system.
\newblock Anthropic Engineering Blog.
\newblock Published June 13, 2025. Accessed 2025-08-19.

\bibitem[{Haruka(2025)}]{Haruka2025}
Haruka, Y. 2025.
\newblock Japan’s Defense AI Roadmap.
\newblock \emph{Nikkei Asian Review}.
\newblock Accessed 2025-07-07.

\bibitem[{Hendrycks, Schmidt, and Wang(2025)}]{Hendrycks2025}
Hendrycks, D.; Schmidt, E.; and Wang, A. 2025.
\newblock Superintelligence Strategy: Expert Version.
\newblock arXiv:2503.05628.

\bibitem[{Hindy(2023)}]{Hindy2023}
Hindy, B. 2023.
\newblock Intelligence Community Building Its Own GPT for Classified Data.
\newblock Accessed 2025-07-07.

\bibitem[{{IARPA}(2021)}]{IARPA2021}
{IARPA}. 2021.
\newblock SMART Program Broad Agency Announcement.
\newblock Technical report.
\newblock Accessed 2025-07-07.

\bibitem[{{IARPA}(2022)}]{IARPA2022}
{IARPA}. 2022.
\newblock Space-Based Machine Automated Recognition Technique (SMART) Program.
\newblock Technical report.
\newblock Accessed 2025-07-07.

\bibitem[{Insights(2020)}]{Deloitte2020}
Insights, D. 2020.
\newblock The Future of Intelligence Analysis in an AI World.
\newblock Accessed 2025-07-07.

\bibitem[{Institute(2024)}]{LibertadND}
Institute, L. 2024.
\newblock AI and Intelligence in Latin America: A Primer.
\newblock White Paper.

\bibitem[{{Iran International}(2025)}]{IranIntl2025}
{Iran International}. 2025.
\newblock Iran’s Revolutionary Guard Tests Domestic AI Surveillance.
\newblock Accessed 2025-07-07.

\bibitem[{Jahankhani et~al.(2020)}]{Jahankhani2020}
Jahankhani, H.; et~al. 2020.
\newblock \emph{Intelligence and Security Informatics: Techniques and Applications}.
\newblock Springer.

\bibitem[{Jensen(2023)}]{Jensen2023}
Jensen, B. 2023.
\newblock From OODA to LLM: Decision Advantage in the Age of Generative AI.
\newblock \emph{Parameters}, 53(4): 15--29.

\bibitem[{Jin et~al.(2025)Jin, Zeng, Yue, Yoon, Arik, Wang, Zamani, and Han}]{jin2025searchr1}
Jin, B.; Zeng, H.; Yue, Z.; Yoon, J.; Arik, S.~O.; Wang, D.; Zamani, H.; and Han, J. 2025.
\newblock Search-R1: Training LLMs to Reason and Leverage Search Engines with Reinforcement Learning.
\newblock \emph{arXiv preprint arXiv:2503.09516}.

\bibitem[{Johnson(2021)}]{Johnson2021}
Johnson, L.~K. 2021.
\newblock National security by numbers: Technology and the intelligence community.
\newblock \emph{Intelligence and National Security}, 36(1): 1--23.

\bibitem[{Jones(2023)}]{jones2023social}
Jones, J.~M. 2023.
\newblock Social Media Users More Inclined to Browse Than Post Content.
\newblock \emph{Gallup}.
\newblock Accessed 2025-07-25.

\bibitem[{Kemp(2025)}]{kemp2025digital}
Kemp, S. 2025.
\newblock Digital 2025: Global Overview Report.
\newblock Accessed 2025-07-25.

\bibitem[{Koch(2024{\natexlab{a}})}]{Koch2024}
Koch, B. 2024{\natexlab{a}}.
\newblock Beyond Human Analysis: AI and the Future of Intelligence Fusion.
\newblock \emph{Journal of Defense Analytics}, 2(1): 15--31.

\bibitem[{Koch(2024{\natexlab{b}})}]{Koch2024b}
Koch, B. 2024{\natexlab{b}}.
\newblock Inference Bottlenecks in Military AI Systems.
\newblock \emph{Defense AI Review}, 3(1): 67--83.

\bibitem[{Kreps(2021)}]{Kreps2021NonstateAI}
Kreps, S. 2021.
\newblock Democratizing Harm: Artificial Intelligence in the Hands of Nonstate Actors.
\newblock Technical report, Brookings Institution.

\bibitem[{Labs(2022)}]{GolemND}
Labs, G. 2022.
\newblock Open-Source OSINT Toolkit.
\newblock GitHub Repository.
\newblock Accessed 2025-07-07.

\bibitem[{Lavoy(2024)}]{RAND_2024_BulkDataAI}
Lavoy, N. 2024.
\newblock Addressing the National Security Risks of Bulk Data in the Age of AI.
\newblock RAND Corporation Commentary.

\bibitem[{Liu et~al.(2023)Liu, Iter, Xu, Wang, Xu, and Zhu}]{liu-etal-2023-g}
Liu, Y.; Iter, D.; Xu, Y.; Wang, S.; Xu, R.; and Zhu, C. 2023.
\newblock {G}-Eval: {NLG} Evaluation using Gpt-4 with Better Human Alignment.
\newblock In Bouamor, H.; Pino, J.; and Bali, K., eds., \emph{Proceedings of the 2023 Conference on Empirical Methods in Natural Language Processing}, 2511--2522. Singapore: Association for Computational Linguistics.

\bibitem[{Martin and Manson(2023)}]{MartinManson2023}
Martin, P.; and Manson, K. 2023.
\newblock CIA to Launch ChatGPT-Style Tool for Analysts.
\newblock Accessed 2025-07-07.

\bibitem[{Maslej et~al.(2025)}]{Maslej2025}
Maslej, N.; et~al. 2025.
\newblock 2025 AI Index Report.
\newblock Accessed 2025-07-07.

\bibitem[{McMahon(2024)}]{McMahon2024}
McMahon, R. 2024.
\newblock Integrating AI into Joint Operations: Lessons Learned.
\newblock \emph{Joint Force Quarterly}, (113): 12--20.

\bibitem[{Mitchell(2025)}]{Mitchell2025GPT4oTopSecret}
Mitchell, B. 2025.
\newblock OpenAI's GPT-4o gets green light for top secret use in Microsoft's Azure cloud.
\newblock DefenseScoop.
\newblock Reports authorization for multimodal GPT-4o in the Azure Government Top Secret cloud; references prior isolated, air-gapped deployment. Accessed: 2025-08-19.

\bibitem[{Mohsin(2024)}]{Mohsin2024}
Mohsin, S. 2024.
\newblock Google Cloud’s AI Tools Boost Maven Phase II.
\newblock \emph{Bloomberg}.
\newblock Accessed 2025-07-07.

\bibitem[{Moran(2023)}]{Moran2023}
Moran, M. 2023.
\newblock Bridging the Gap: Implementing AI Across the US Intelligence Community.
\newblock \emph{Studies in Intelligence}, 67(3): 45--62.

\bibitem[{Nakano et~al.(2021)Nakano, Hilton, Balaji, Wu, Ouyang, Kim, Hesse, Jain, Kosaraju, Saunders, Jiang, Cobbe, Eloundou, Krueger, Button, Knight, Chess, and Schulman}]{nakano2021webgpt}
Nakano, R.; Hilton, J.; Balaji, S.; Wu, J.; Ouyang, L.; Kim, C.; Hesse, C.; Jain, S.; Kosaraju, V.; Saunders, W.; Jiang, X.; Cobbe, K.; Eloundou, T.; Krueger, G.; Button, K.; Knight, M.; Chess, B.; and Schulman, J. 2021.
\newblock WebGPT: Browser-assisted question-answering with human feedback.
\newblock \emph{arXiv preprint arXiv:2112.09332}.

\bibitem[{{National Academies of Sciences, Engineering, and Medicine}(2025)}]{NASEM2025}
{National Academies of Sciences, Engineering, and Medicine}. 2025.
\newblock \emph{The Age of {AI} in the Life Sciences: Benefits and Biosecurity Considerations}.
\newblock Washington, DC: The National Academies Press.

\bibitem[{{National Institute of Standards and Technology}(2023)}]{NIST_AIRMF_1_0_2023}
{National Institute of Standards and Technology}. 2023.
\newblock Artificial Intelligence Risk Management Framework (AI RMF 1.0).
\newblock Technical Report NIST AI 100-1, NIST.

\bibitem[{{National Security Commission on Artificial Intelligence}(2021{\natexlab{a}})}]{NSCAI2021}
{National Security Commission on Artificial Intelligence}. 2021{\natexlab{a}}.
\newblock Final Report.
\newblock Accessed 2025-07-07.

\bibitem[{{National Security Commission on Artificial Intelligence}(2021{\natexlab{b}})}]{NSCAI2021FinalReport}
{National Security Commission on Artificial Intelligence}. 2021{\natexlab{b}}.
\newblock Final Report.

\bibitem[{{NATO}(1996)}]{NATO1996}
{NATO}. 1996.
\newblock Allied Joint Doctrine for Intelligence, Surveillance, and Reconnaissance.

\bibitem[{Nelson and Epstein(2022)}]{NelsonEpstein2022}
Nelson, Z.; and Epstein, J. 2022.
\newblock ChatBIT: A Chinese Large Language Model for Intelligence Analysis.
\newblock Technical report, Defense Innovation Unit.
\newblock Accessed 2025-07-07.

\bibitem[{{North Atlantic Treaty Organization (NATO)}(2024)}]{NATO_2024_AI_Strategy_Revised}
{North Atlantic Treaty Organization (NATO)}. 2024.
\newblock Summary of NATO's Revised Artificial Intelligence (AI) Strategy.
\newblock Sets out Principles of Responsible Use for AI in defence and updated cooperation mechanisms.

\bibitem[{Nyhan(2024)}]{Nyhan2024}
Nyhan, B. 2024.
\newblock Multimodal Misinformation: Risks and Mitigations.
\newblock \emph{Science}, 384(6592): 46--49.

\bibitem[{{Office of the Director of National Intelligence}(2024)}]{ODNI2024}
{Office of the Director of National Intelligence}. 2024.
\newblock Annual Threat Assessment of the US Intelligence Community.
\newblock Accessed 2025-07-07.

\bibitem[{{OpenAI}()}]{openai2025o3}
{OpenAI}. ????
\newblock Introducing OpenAI o3 and o4-mini.

\bibitem[{{OpenAI}(2025)}]{deepresearch}
{OpenAI}. 2025.
\newblock Introducing Deep Research.
\newblock Blog post.

\bibitem[{OpenAI(2025)}]{OpenAIpricing}
OpenAI. 2025.
\newblock OpenAI API Pricing.
\newblock Web Page.
\newblock Accessed 2025-07-07.

\bibitem[{Ouyang et~al.(2022)Ouyang, Jeff W. Y. Wu, Jiang, Almeida, Carroll L. Wainwright, Mishkin, Zhang, Agarwal, Slama, Ray, Schulman, Hilton, Kelton, Miller, Simens, Askell, Welinder, Christiano, Leike, and Lowe}]{ouyang2022training}
Ouyang, L.; Jeff W. Y. Wu; Jiang, X.; Almeida, D.; Carroll L. Wainwright; Mishkin, P.; Zhang, C.; Agarwal, S.; Slama, K.; Ray, A.; Schulman, J.; Hilton, J.; Kelton, F.; Miller, L.; Simens, M.; Askell, A.; Welinder, P.; Christiano, P.; Leike, J.; and Lowe, R. 2022.
\newblock Training Language Models to Follow Instructions with Human Feedback.
\newblock In \emph{Advances in Neural Information Processing Systems 35}.
\newblock ArXiv:2203.02155.

\bibitem[{Ownby and Kott(2006)}]{ownby_kott_2006_raid}
Ownby, M.; and Kott, A. 2006.
\newblock Reading the Mind of the Enemy: Predictive Analysis and Command Effectiveness.
\newblock In \emph{Proceedings of the 2006 Command and Control Research and Technology Symposium (CCRTS): The State of the Art and the State of the Practice}. San Diego, CA.
\newblock A version appears as arXiv:1607.06759.

\bibitem[{Panella(2024)}]{Panella2024}
Panella, J. 2024.
\newblock Detecting Illegal Fishing with New Satellite Data.
\newblock Accessed 2025-07-07.

\bibitem[{Park, Lee, and Ko(2022)}]{park2022}
Park, J.; Lee, B.; and Ko, E. 2022.
\newblock A comprehensive study on anomaly score for GAN-based anomaly detection in surveillance videos.
\newblock \emph{IEEE Transactions on Information Forensics and Security}, 17: 2879--2891.

\bibitem[{Pellerin(2017)}]{Pellerin2017}
Pellerin, C. 2017.
\newblock Project Maven to Deploy Computer Vision Algorithms for Warzone Intelligence.
\newblock Accessed 2025-07-07.

\bibitem[{Pomfret and Pang(2024)}]{PomfretPang2024}
Pomfret, J.; and Pang, V. 2024.
\newblock Inside Beijing’s Quest for AI-Enhanced Intelligence.
\newblock Accessed 2025-07-07.

\bibitem[{Ratcliffe(2024)}]{Ratcliffe2024}
Ratcliffe, R. 2024.
\newblock OSINT Analysts Track Myanmar Military Plane Movements.
\newblock Accessed 2025-07-07.

\bibitem[{Rising and Tang(2024)}]{RisingTang2024}
Rising, D.; and Tang, D. 2024.
\newblock China’s Prototype Reactor for Nuclear-Powered Aircraft Carriers Revealed by Satellite Images.
\newblock Accessed 2025-07-07.

\bibitem[{Rivera et~al.(2024)Rivera, Mukobi, Reuel, Lamparth, Smith, and Schneider}]{Rivera2024}
Rivera, J.-P.; Mukobi, G.; Reuel, A.; Lamparth, M.; Smith, C.; and Schneider, J. 2024.
\newblock Escalation Risks from Language Models in Military and Diplomatic Decision-Making.
\newblock \emph{Proceedings of the 2024 ACM Conference on Fairness, Accountability, and Transparency}, 63.

\bibitem[{Rose et~al.(2020)Rose, Borchert, Mitchell, and Connelly}]{NIST_SP_800_207_2020}
Rose, S.; Borchert, O.; Mitchell, S.; and Connelly, S. 2020.
\newblock Zero Trust Architecture.
\newblock Technical Report NIST Special Publication 800-207, National Institute of Standards and Technology.

\bibitem[{Salisbury(2024)}]{Salisbury2024}
Salisbury, P. 2024.
\newblock North Korea’s Sanctions-Evasion Shipping Routes.
\newblock Accessed 2025-07-07.

\bibitem[{Sayler(2022)}]{CRS2022}
Sayler, K.~M. 2022.
\newblock Joint All-Domain Command and Control (JADC2).
\newblock Accessed 2025-07-07.

\bibitem[{Schick et~al.(2023)Schick, Dwivedi{-}Yu, Dessì, Raileanu, Lomeli, Zettlemoyer, Cancedda, and Scialom}]{schick2023toolformer}
Schick, T.; Dwivedi{-}Yu, J.; Dessì, R.; Raileanu, R.; Lomeli, M.; Zettlemoyer, L.; Cancedda, N.; and Scialom, T. 2023.
\newblock Toolformer: Language Models Can Teach Themselves to Use Tools.
\newblock In \emph{Advances in Neural Information Processing Systems 36}.
\newblock ArXiv:2302.04761.

\bibitem[{Schroeder et~al.(2025)Schroeder, Morgan, Luo, and Glass}]{schroeder-etal-2025-thread}
Schroeder, P.; Morgan, N.~W.; Luo, H.; and Glass, J.~R. 2025.
\newblock {THREAD}: Thinking Deeper with Recursive Spawning.
\newblock In \emph{Proceedings of the 2025 Conference of the Nations of the Americas Chapter of the Association for Computational Linguistics: Human Language Technologies (Volume 1: Long Papers)}, 8418--8442. Albuquerque, New Mexico: Association for Computational Linguistics.
\newblock ISBN 979-8-89176-189-6.

\bibitem[{Sharma(2025)}]{Sharma2025}
Sharma, K. 2025.
\newblock Low-Latency Inference on Edge GPUs.
\newblock \emph{IEEE Micro}, 45(2): 50--63.

\bibitem[{Sheldon(2024)}]{Sheldon2024}
Sheldon, J. 2024.
\newblock Missile Strike on Kyiv’s Okhmatdyt Hospital Geolocated by OSINT Analysts.
\newblock Accessed 2025-07-07.

\bibitem[{Simmons and Edler(2024)}]{SimmonsEdler2024}
Simmons, J.; and Edler, M. 2024.
\newblock Autonomous Systems and Strategic Stability.
\newblock \emph{Survival}, 66(2): 55--78.

\bibitem[{Soldatov and Borogan(2011)}]{SoldatovBorogan2011}
Soldatov, A.; and Borogan, I. 2011.
\newblock \emph{The New Nobility: The Restoration of Russia’s Security State and the Enduring Legacy of the KGB}.
\newblock PublicAffairs.

\bibitem[{SpecialEurasia(2025)}]{SpecialEurasia2025}
SpecialEurasia. 2025.
\newblock North Korea’s AI-Powered Reconnaissance Satellites.
\newblock \emph{SpecialEurasia Intelligence Brief}, 5(1).

\bibitem[{Strout(2022)}]{Strout2022}
Strout, N. 2022.
\newblock US Army Accelerates Project Maven AI Tools.
\newblock Accessed 2025-07-07.

\bibitem[{{Sustainability Directory}(2025)}]{sustainability_directory_ai_esg_2025}
{Sustainability Directory}. 2025.
\newblock Could {AI} Trends Suggest Future {ESG} Access? {AI} trends indicate enhanced access to {ESG} data, offering deeper insights for corporate reporting and sustainable investment decisions.
\newblock Accessed 2025-07-25.

\bibitem[{Tetlock and Gardner(2015)}]{tetlock2015superforecasting}
Tetlock, P.~E.; and Gardner, D. 2015.
\newblock \emph{Superforecasting: The Art and Science of Prediction}.
\newblock New York, NY: Crown.
\newblock ISBN 978-0804136716.

\bibitem[{{The White House}(2022)}]{WhiteHouse_CHIPS_FactSheet_2022}
{The White House}. 2022.
\newblock FACT SHEET: CHIPS and Science Act Will Lower Costs, Create Jobs, Strengthen Supply Chains, and Counter China.

\bibitem[{Toniolo et~al.(2023)}]{Toniolo2023}
Toniolo, A.; et~al. 2023.
\newblock Human–AI Collaboration in Intelligence Analysis.
\newblock \emph{Computers in Human Behavior}, 139: 107--123.

\bibitem[{Toreini et~al.(2020)Toreini, Aitken, Coopamootoo, Elliott, Zelaya, and Van~Moorsel}]{toreini2020}
Toreini, E.; Aitken, M.; Coopamootoo, K.; Elliott, K.; Zelaya, C.~G.; and Van~Moorsel, A. 2020.
\newblock The relationship between trust in AI and trustworthy machine learning technologies.
\newblock \emph{Proceedings of the 2020 conference on fairness, accountability, and transparency}, 272--283.

\bibitem[{Tsanousa(2022)}]{Tsanousa2022}
Tsanousa, M. 2022.
\newblock Small Satellites, Big Data: Managing Proliferating GEOINT Feeds.
\newblock \emph{Space Policy}, 61: 101--115.

\bibitem[{Tsvetovat and Carley(2003)}]{tsvetovat2003structural}
Tsvetovat, M.; and Carley, K.~M. 2003.
\newblock Structural Knowledge and Success of Anti-Terrorist Activity.
\newblock \emph{Journal of Social Structure}, 6.
\newblock Supported by Department of Defense, Office of Naval Research Grant No. 9620.1.1140071, NSF IRI9633 662 and NSF IGERT 9972762.

\bibitem[{{UK Government Communications Headquarters}(2021)}]{GCHQ2021}
{UK Government Communications Headquarters}. 2021.
\newblock Pioneering a New National Cyber AI Hub.
\newblock Accessed 2025-07-07.

\bibitem[{{UK Ministry of Defence}(2024)}]{UKMOD2024}
{UK Ministry of Defence}. 2024.
\newblock Defence Artificial Intelligence Strategy.
\newblock Accessed 2025-07-07.

\bibitem[{{UNICRI} and {UNCCT}(2021)}]{UNICRI2021AlgorithmsTerrorism}
{UNICRI}; and {UNCCT}. 2021.
\newblock Algorithms and Terrorism: The Malicious Use of Artificial Intelligence for Terrorist Purposes.

\bibitem[{{U.S. Department of Commerce, Bureau of Industry and Security}(2023)}]{BIS_FedReg_AdvancedComputing_2023}
{U.S. Department of Commerce, Bureau of Industry and Security}. 2023.
\newblock Implementation of Additional Export Controls: Certain Advanced Computing Items; Supercomputer and Semiconductor End Use; Updates and Corrections.
\newblock Federal Register, 88 FR 73458.

\bibitem[{{U.S. Department of Justice, National Security Division}(2025)}]{DOJ2025_EO14117_FinalRule}
{U.S. Department of Justice, National Security Division}. 2025.
\newblock Preventing Access to U.S. Sensitive Personal Data and Government-Related Data by Countries of Concern or Covered Persons.
\newblock Final rule, \emph{Federal Register}, 28 CFR Part 202.
\newblock Implements Executive Order 14117; effective April 8, 2025.

\bibitem[{{US DHS}(2023)}]{DHS2023}
{US DHS}. 2023.
\newblock Artificial Intelligence in Homeland Security: FY2023 Progress.
\newblock Accessed 2025-07-07.

\bibitem[{{US DOD}(2024)}]{DOD2024}
{US DOD}. 2024.
\newblock 2024 Defense Intelligence Strategy.
\newblock Accessed 2025-07-07.

\bibitem[{{US Intelligence Community}(2024)}]{USIC2024}
{US Intelligence Community}. 2024.
\newblock Vision for the Intelligence Community 2024.
\newblock Accessed 2025-07-07.

\bibitem[{Ustin and Middleton(2024)}]{ustin2024earth}
Ustin, S.~L.; and Middleton, E.~M. 2024.
\newblock Current and Near-Term Earth-Observing Environmental Satellites, Their Missions, Characteristics, Instruments, and Applications.
\newblock \emph{Sensors}, 24(11): 3488.

\bibitem[{Vassilev et~al.(2025)Vassilev, Oprea, Fordyce, Anderson, Davies, and Hamin}]{NIST_AI_100_2e2025}
Vassilev, A.; Oprea, A.; Fordyce, A.; Anderson, H.; Davies, X.; and Hamin, M. 2025.
\newblock Adversarial Machine Learning: A Taxonomy and Terminology of Attacks and Mitigations.
\newblock Technical Report NIST AI 100-2e2025, National Institute of Standards and Technology.

\bibitem[{Wei(2018)}]{Wei2018}
Wei, L. 2018.
\newblock Crowdsourcing Analysis: Leveraging Public Data for Intelligence.
\newblock \emph{International Journal of Intelligence}, 9(1): 1--18.

\bibitem[{Weiner and Short(n.d.)}]{WeinerShortND}
Weiner, T.; and Short, B. n.d.
\newblock Signals Intelligence Collection: A Historical Overview.
\newblock Technical report, National Security Archive.
\newblock Accessed 2025-07-07.

\bibitem[{Welch(2025)}]{Welch2025}
Welch, B. 2025.
\newblock NATO Turns to Planet Labs for AI-Enhanced Surveillance.
\newblock Accessed 2025-07-07.

\bibitem[{Whitney(2025{\natexlab{a}})}]{Whitney2025}
Whitney, J. 2025{\natexlab{a}}.
\newblock Assessing Geospatial Intelligence in 2025: A New Paradigm.

\bibitem[{Whitney(2025{\natexlab{b}})}]{Whitney2025b}
Whitney, J. 2025{\natexlab{b}}.
\newblock Inference at the Edge: Deploying AI in Contested Environments.

\bibitem[{Wirtschafter(2024)}]{Wirtschafter2024}
Wirtschafter, J. 2024.
\newblock Israel Tests AI-Driven Targeting System in Gaza.
\newblock Accessed 2025-07-07.

\bibitem[{Wong(2023)}]{Wong2023}
Wong, E. 2023.
\newblock China Uses AI to Identify US Intelligence Operatives.
\newblock Accessed 2025-07-07.

\bibitem[{Yue et~al.(2023)}]{Yue2023}
Yue, X.; et~al. 2023.
\newblock Vision-Language Models Are Multimodal Few-Shot Learners.
\newblock \emph{Advances in Neural Information Processing Systems}.

\bibitem[{Zhang, Yuan, and Yao(2023)}]{zhang2023meta}
Zhang, Y.; Yuan, Y.; and Yao, A. C.-C. 2023.
\newblock Meta Prompting for {AI} Systems.
\newblock \emph{arXiv preprint arXiv:2311.11482}.

\bibitem[{Zhu, Dugan, and Callison-Burch(2024)}]{zhu-etal-2024-redel}
Zhu, A.; Dugan, L.; and Callison-Burch, C. 2024.
\newblock {R}e{D}el: A Toolkit for {LLM}-Powered Recursive Multi-Agent Systems.
\newblock In \emph{Proceedings of the 2024 Conference on Empirical Methods in Natural Language Processing: System Demonstrations}, 162--171. Miami, Florida, USA: Association for Computational Linguistics.

\end{thebibliography}

\newpage

\end{document}